%% file: acl_latex.tex
\documentclass[11pt]{article}

\usepackage[final]{acl}

\usepackage{times}
\usepackage{latexsym}

\usepackage[T1]{fontenc}

\usepackage[utf8]{inputenc}
\usepackage{enumitem}

\usepackage{microtype}

\usepackage{inconsolata}

\usepackage{graphicx}
\usepackage{amsfonts}
\usepackage{amssymb}
\usepackage{algorithm}
\usepackage{algpseudocode}
\usepackage{booktabs}
\usepackage{array}
\usepackage{multirow}
\usepackage{graphicx}
\usepackage[most]{tcolorbox}
\usepackage{fontawesome5}
\newcommand{\method}{OCM}

\title{Object-Centric Environment Modeling for Agentic Tasks}

\author{
 \textbf{Yiyang Li\textsuperscript{1}},
 \textbf{Tianyi Ma\textsuperscript{1}},
 \textbf{Zehong Wang\textsuperscript{1}},
 \textbf{Yijun Ma\textsuperscript{1}},
 \textbf{Yanfang Ye\textsuperscript{1\textdagger}}
\\
 \textsuperscript{1}University of Notre Dame \quad \textsuperscript{\textdagger}Corresponding Author \\
 \texttt{\{yli62, yye7\}@nd.edu} \quad \faGithub\ \href{https://github.com/Yiyang-Ian-Li/OCM}{Code}
}

\begin{document}
\maketitle
\begin{abstract}
Large language model (LLM) agents can improve through accumulated experience, but free-form textual memories become difficult to maintain, validate, and reuse as interactions grow. Recent symbolic approaches learn executable skills or programmatic world models, yet often store local procedures or assume simplified dynamics. We propose \textbf{O}bject-\textbf{C}entric Environment \textbf{M}odeling (\textbf{\method{}}), which organizes experience into an executable object-centric environment model. \method{} maintains two connected code bases: object knowledge, which defines environment entities and mechanisms as Python classes, and procedure knowledge, which records reusable interaction patterns that must import and use the object model. 
OCM works in an online setting: after each episode, \method{} reflects on the trajectory, updates both knowledge bases, and verifies that all procedures execute against the updated object model. During future interaction, the agent uses progressive knowledge disclosure to inspect compact code signatures first and read source code only when needed. Experiments show that \method{} achieves the best average rank across benchmarks and reduces invalid actions, demonstrating that agents can benefit from building object-centric environment models.
\end{abstract}

\section{Introduction}

\begin{figure}[!t]
    \centering
    \includegraphics[width=\linewidth]{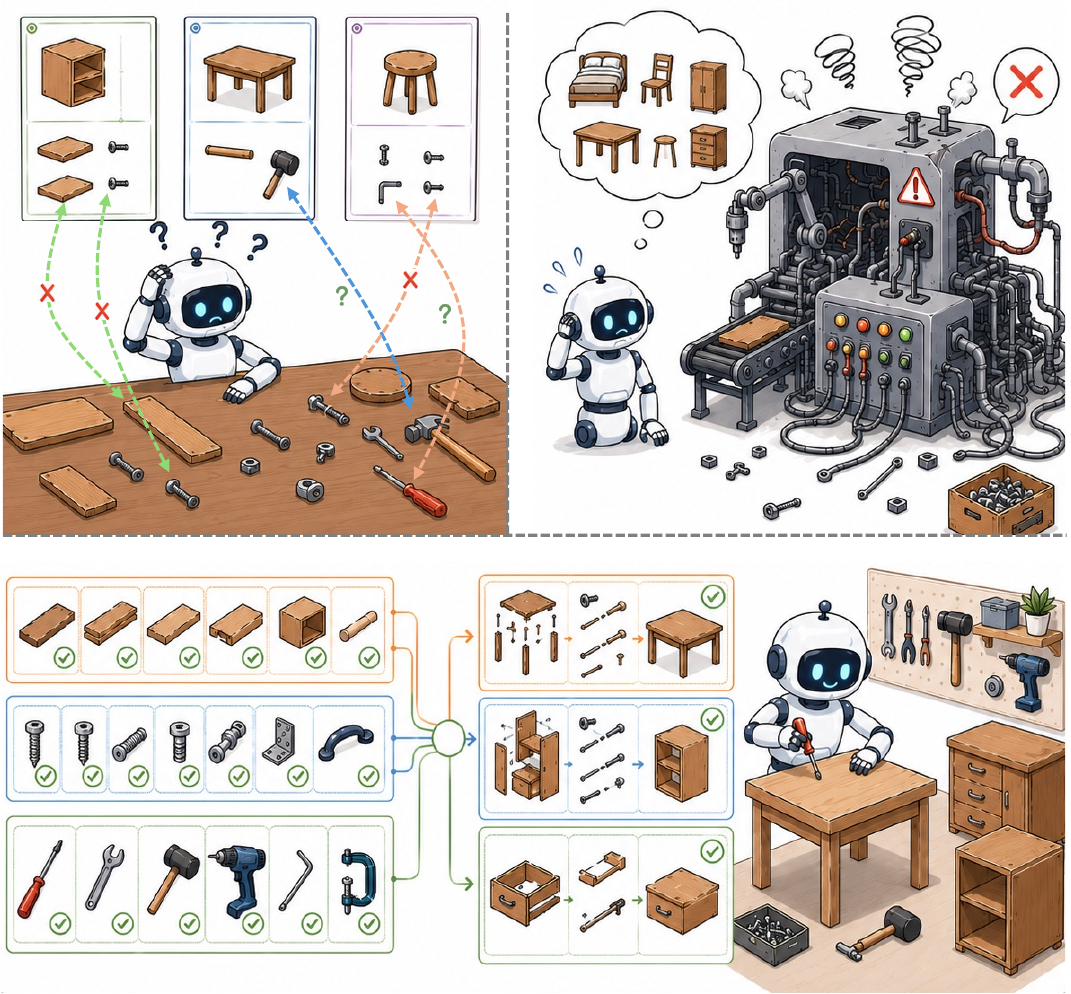}
    \caption{
    \textbf{Upper left}: Isolated skills resemble scattered manuals whose assumed tools and parts are potentially inconsistent.
    \textbf{Upper right}: Monolithic symbolic world models resemble an overcomplicated machine that attempts to model all transitions.
    \textbf{Bottom}: \method{} standardizes objects and mechanisms first, then grounds procedural knowledge in this shared object model, making experience easier to organize, verify, and reuse.
    }
    \label{fig:motivation}
    \vspace{-10pt}
\end{figure}

As large language model (LLM) agents are increasingly deployed in interactive environments, their ability to learn from past experiences has become critical for solving long-horizon and open-ended tasks~\cite{zhang2025survey, hu2025memory,ye2025llms4all,chen2025obvious,chen2025clear}. To enable agents to improve over time, prior work has introduced memory mechanisms that store and reuse previous experience as textual memory streams, verbal reflections, or reusable workflows~\cite{zhao2024expel, wang2024agent, zhong2024memorybank}. Despite their effectiveness, textual knowledge becomes increasingly difficult to manage as agents accumulate more interactions. Newly written knowledge may be redundant or inconsistent with previous ones, while large collections of textual knowledge are hard to validate or audit~\cite{hu2025evaluating, chao2026stale}. These limitations can make agent behavior less controllable and less stable, especially in complex environments where errors in memory may propagate across future decisions~\cite{xiong2025memory, lam2026governing}.

To make accumulated experience easier to inspect and reuse, a growing line of work represents agent knowledge in symbolic or executable forms~\cite{ning2026code}. These methods learn symbolic rules, code-based skills, or programmatic world models, improving interpretability and enabling execution-based validation~\cite{wang2023voyager, wang2025inducing, zheng2026code2world}. However, existing approaches typically capture only one side of the desired structure. Code-based skill learning stores reusable procedures, but these procedures often remain local, without a shared account of the objects, states, affordances, and preconditions that make them valid~\cite{wang2023voyager, zhou2024wall, wang2025inducing}. Programmatic world-modeling methods instead aim to model environment dynamics explicitly, but are usually studied in settings with smaller action spaces and well-specified states~\cite{tang2024worldcoder, khan2025one, piriyakulkij2026poe}. This leaves a missing middle ground: agents need symbolic knowledge that is more coherent than isolated skills, but less demanding than a monolithic transition model. Figure~\ref{fig:motivation} summarizes this tension. The key challenge is to organize experience around persistent environment entities and mechanisms, so that procedural knowledge can be reused while remaining consistent with a shared object-level model.

To address this gap, we propose \textbf{O}bject-\textbf{C}entric Environment \textbf{M}odeling (\textbf{\method{}}), a framework that enables an agent to construct and maintain an object-centric model of its environment through interaction. \method{} represents the agent's knowledge with two mutually connected code bases. The first, \emph{object knowledge}, defines Python classes for the objects in the environment; class attributes and methods encode object properties and interaction mechanisms. The second, \emph{procedure knowledge}, stores interaction experience as executable Python files. Each procedure file describes a concrete piece of knowledge or experience, and must import and use the classes defined in object knowledge. After each interaction episode, the agent updates both forms of knowledge and executes the procedure files to check whether the accumulated knowledge remains mutually consistent, turning past experience into an auditable, executable, and object-grounded environment model. 
During future interactions, the \method{} agent can inspect the signatures of object classes and procedure functions and selectively read their source code. \method{} encourages the agent to organize knowledge around persistent entities and their interaction mechanisms, enabling more structured knowledge accumulation, self-consistency checking, and more controllable reuse of prior experience in complex environments.

We evaluate \method{} on three interactive benchmarks: ScienceWorld, ALFWorld, and PlanCraft. OCM achieves the best average rank among compared methods and reduces invalid actions. These results indicate that grounding procedural experience in a shared object model helps agents accumulate reusable environment knowledge and act more reliably in complex interactive tasks. Our main contributions are as follows:

\begin{itemize}[leftmargin=*]
    \item We propose \textbf{object-centric environment modeling}, a framework that represents agent experience as an executable environment model composed of object knowledge and object-grounded procedure knowledge.
    \item We introduce a continual learning loop that reflects on completed trajectories, updates the object-centric code base, and verifies executable consistency before committing new knowledge.
    \item We validate \method{} on three interactive benchmarks and analyze how its learned knowledge improves performance and action feasibility.
\end{itemize}

\section{Related Work}

\subsection{Memory-Augmented Agents}

Memory mechanisms enable LLM agents to reuse past experience across long-horizon interactions~\cite{kagaya2024rap, chen2024automanual, zhang2025survey, hu2025memory}. Most existing methods store experience primarily as natural-language memories: Reflexion stores verbal feedback from previous trials to improve later attempts~\cite{shinn2023reflexion}; ExpeL extracts reusable lessons from successful and failed trajectories~\cite{zhao2024expel}; and Agent Workflow Memory abstracts trajectories into reusable workflows~\cite{wang2024agent}. These methods show that experience reuse can substantially improve agent behavior. But free-form textual memory is difficult to maintain as it grows, learned memories may become stale, redundant, or mutually inconsistent, and large memory collections are hard to validate or audit~\cite{hu2025evaluating, chao2026stale, xiong2025memory, lam2026governing}. \method{} shares the goal of continual experience accumulation, but represents experience as executable object-grounded code rather than only textual memory.

\subsection{Symbolic World Modeling}

Symbolic and executable representations provide a more structured substrate for agent knowledge. Prior work in object-oriented modeling represents environments through objects, attributes, relations, and transition mechanisms, enabling abstraction and transfer~\cite{diuk2008object, kansky2017schema}. Recent LLM agent methods similarly use symbolic programs to store reusable knowledge~\cite{dainese2024generating, shao2026lifting, zhao2026neuro}. For example, Voyager, Wall-E, and ASI accumulate executable skills in code~\cite{wang2023voyager, zhou2024wall, wang2025inducing}; WorldCoder and PoE-World learn Python world models for planning~\cite{tang2024worldcoder, piriyakulkij2026poe}. These approaches improve interpretability and enable execution-based verification, but they typically focus on either local skills or world modeling in structured environments. In contrast, \method{} couples procedural experience with a shared object-centric code model that defines persistent entities and mechanisms. This encourages a unified, auditable, and self-consistent environment model that can be incrementally updated through interaction.

\begin{figure*}[!t]
    \centering
    \includegraphics[width=\linewidth]{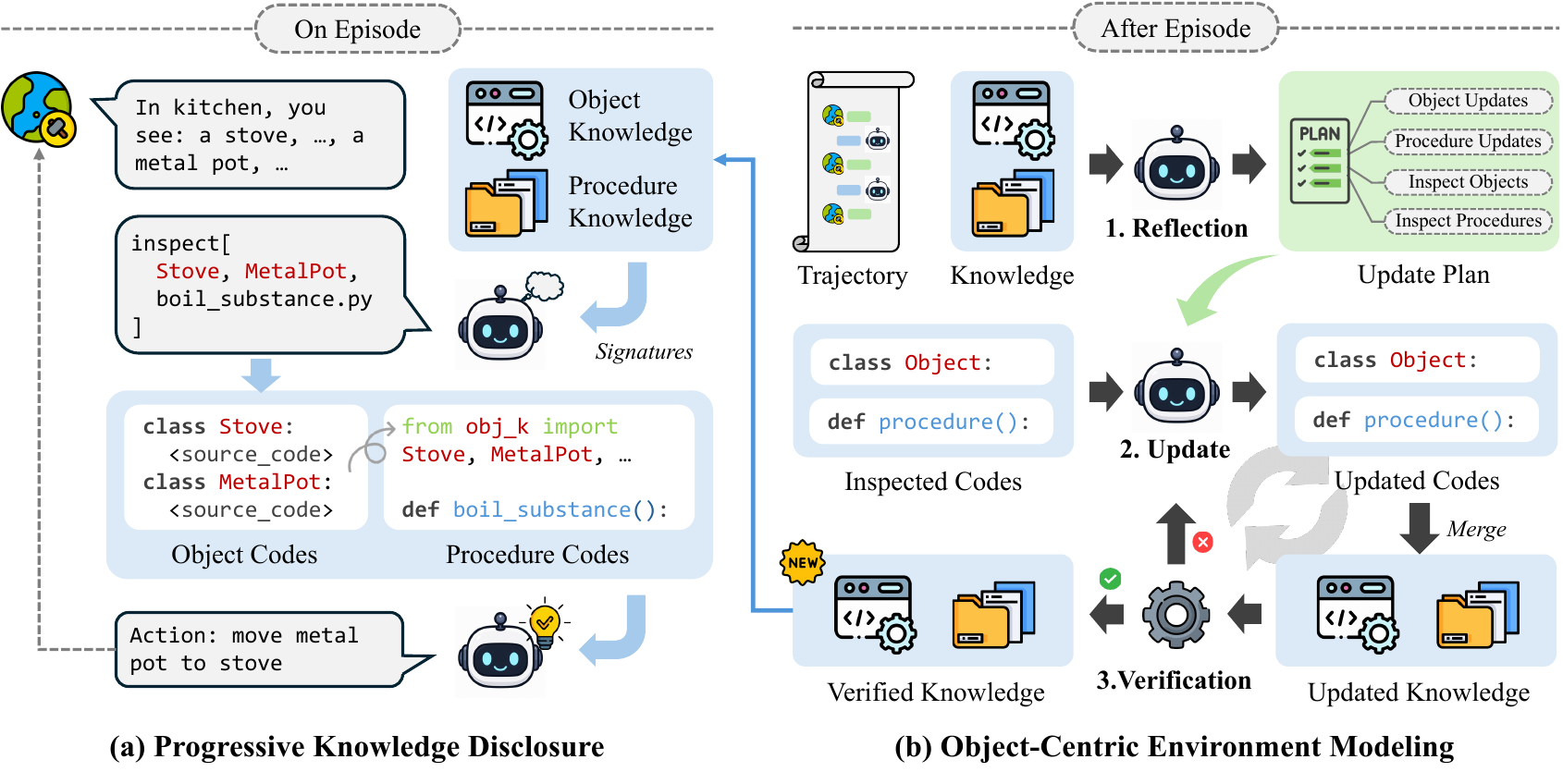}
    \caption{
        \textbf{Overview of \method{}.} 
        (a) During an episode, the agent uses progressive knowledge disclosure: it first observes signatures of learned object and procedure codes, then selectively inspects relevant source code before choosing an environment action. 
        (b) After each episode, \method{} converts the trajectory into an object-centric update plan, generates new candidate object and procedure code, and commits the update only after verification, where all procedure files must execute successfully against the updated object code. By grounding procedure knowledge in shared object knowledge, \method{} maintains an executable and auditable environment model for future interactions.
        }
    \label{fig:method}
    \vspace{-10pt}
\end{figure*}

\section{Object-Centric Environment Modeling}

\subsection{Preliminary}

\paragraph{Problem Formulation.}
We consider a partially observable interactive environment with observation space $\mathcal{O}$ and action space $\mathcal{A}$. An agent continually interacts with the environment over a sequence of $K$ task episodes $\{\tau_k\}_{k=1}^K$. At step $t$ in episode $k$, the agent observes $o_t^k \in \mathcal{O}$ and selects an action $a_t^k \in \mathcal{A}$. The trajectory of episode $k$ is denoted as
$H^k = (o_0^k, a_1^k, o_1^k, \ldots, a_T^k, o_T^k),$
where $o_0^k$ includes the initial task description. After the episode terminates, the agent receives a final reward $R^k \in [0,1]$. The goal is to maximize task success and cumulative reward $\sum_{k=1}^K R^k$ through continual interaction and knowledge accumulation.

\noindent \paragraph{Knowledge Representation.}
\method{} maintains an executable environment model consisting of two connected knowledge bases. The first is \emph{object knowledge}, denoted $K_o$, implemented as a Python module that defines object classes and helper functions. These classes encode reusable entities in the environment, including their observable states, affordances, constraints, relations, and transition mechanisms. The second is \emph{procedure knowledge}, denoted $K_p$, implemented as a set of Python files. Each procedural file records one focused interaction pattern or experiential rule, such as when a mechanism applies, why an action succeeds or fails, or how feedback should be interpreted. A central constraint of \method{} is that procedure knowledge must be grounded in object knowledge. A procedural file is not allowed to stand as an isolated trajectory summary; it must import and use classes or functions from $K_o$. Thus, episodic experience is linked to persistent object-level abstractions rather than being stored as disconnected textual memories or action scripts. Together, $(K_o, K_p)$ forms the agent's object-centric environment model.

\noindent \paragraph{Overview.}
During an episode $k$, if knowledge has been accumulated, the agent acts using the current environment model. The agent is given compact interfaces of $K_o^k$ and $K_p^k$: signatures of object classes and method functions, together with an index of procedural files and their public functions. These summaries expose the structure of the learned model without placing all accumulated source code into the prompt.

After an episode ends, \method{} updates its environment model through two stages. First, an object-centric reflection stage converts the trajectory into a structured learning plan, identifying which reusable object mechanisms or procedural patterns may need to be created or revised. Second, a knowledge update stage generates concrete Python code for the planned changes and validates the resulting candidate knowledge base. An update is committed only if the candidate code is executable against the current object model.

As illustrated in Figure~\ref{fig:method}, this yields a closed learning loop: future decisions are made from the current object-centric model; new feedback is reflected into the object-grounded model; and only executable updates that remain consistent with prior knowledge are incorporated.

\subsection{Object-Centric Knowledge Reflection}

The reflection stage converts a completed episode into a structured learning plan. Instead of directly rewriting the knowledge base from the raw trajectory, \method{} first decides \emph{what} should be learned. This separation prevents incidental episode details, such as transient locations, repeated invalid actions, or task-specific action sequences, from being stored as long-term knowledge.

Given the trajectory $H^k$, task results $R^k$, the object signatures $\mathrm{Sig}(K_o^k)$, and the procedural index $\mathrm{Ind}(K_p^k)$, the reflection model produces
\[
\begin{aligned}
P^k
&= \mathrm{Reflect}\!\left(H^k, R^k, \mathrm{Sig}(K_o^k), \mathrm{Ind}(K_p^k)\right) \\
&= \left(P_o^k, P_p^k, I_o^k, I_p^k\right).
\end{aligned}
\]
Here, $P_o^k$ and $P_p^k$ specify planned updates to object knowledge and procedure knowledge, while $I_o^k$ and $I_p^k$ specify object symbols and procedural files whose source code should be inspected before editing. The plan is constrained to contain only reusable mechanisms or decision patterns supported by the episode. Object updates should describe recurring environment mechanisms, such as object states, affordances, preconditions, effects, or relations. For example, the agent should learn that closed containers must be opened before their contents are accessible, rather than memorize that a particular cabinet was closed in one episode. Procedure updates, in turn, must be explained through the object model: a procedural pattern should identify which object classes or methods determine when the pattern applies and why it succeeds or fails.

The inspection fields connect new evidence to prior knowledge. If the episode appears to revise an existing class, function, or procedural pattern, \method{} retrieves the corresponding source code before generating an update. The agent therefore sees both the new trajectory evidence and the relevant old implementation, reducing accidental duplication and making revisions local. Reflection thus acts as a routing mechanism: it filters noisy experience into a small set of object-grounded learning targets and identifies the prior code that those targets must remain compatible with.

\subsection{Object-Centric Knowledge Update}

Given the reflection plan, \method{} generates concrete Python code. The agent receives the plan $P^k$, the trajectory $H^k$, task results $R^k$, and the inspected source code $S^k = \mathrm{Src}(I_o^k, I_p^k)$. It outputs the candidate object and procedure code:
\[
(\Delta C_o^k, \Delta C_p^k)
=
\mathrm{Update}(P^k, H^k, R^k, S^k).
\]

Object updates $\Delta C_o^k$ are merged into $K_o^k$. New top-level definitions are appended, while updated definitions replace existing definitions with the same name. This allows the object model to evolve as new evidence accumulates while keeping a single shared namespace for environment entities and mechanisms. The resulting object module is then compiled as part of validation before it can affect future decisions.

Procedural updates $\Delta C_p^k$ are stored as separate Python files. Each procedural file must encode one focused interaction pattern and must import and use relevant symbols from $K_o^k$. A procedure may define workflows, rules, failure patterns, heuristics, or recovery principles. It must not define a complete action controller or replay an entire trajectory. This restriction distinguishes procedure knowledge from executable policies: procedures explain reusable mechanisms and decision patterns, but they do not directly control the agent.

Before committing a candidate update, \method{} verifies that the full procedural knowledge remains executable against the candidate object knowledge:
\[
\mathrm{Verify}(\widetilde{K}_o,\widetilde{K}_p)
=
\bigwedge_{p \in \widetilde{K}_p}
\mathrm{Exec}(p \mid \widetilde{K}_o).
\]
Here, $\mathrm{Exec}(p \mid \widetilde{K}_o)$ returns true only if procedure $p$ executes successfully.
Every procedure file in the current candidate knowledge base $(\widetilde{K}_o,\widetilde{K}_p)$ must execute successfully using the updated object knowledge. If verification fails, the error is returned to the update model for another attempt; otherwise, the candidate is committed.
This verification step does not prove that the learned model is semantically complete or always correct. Instead, it enforces a practical invariant: every committed procedure remains executable against the current object knowledge base. As a result, local update errors are less likely to silently corrupt future decisions, and the accumulated environment model remains auditable as ordinary source code.

\subsection{Progressive Knowledge Disclosure}

The same object-centric structure used for learning also controls how knowledge is exposed during future decision-making. Placing all accumulated code into every prompt would be expensive and distracting. \method{} instead exposes a compact interface first, and reveals source code only when needed.

At step $t$ of episode $k$, the decision context is
\[
C_t^{(0)}
=
\left[
\mathcal{I},
o_t^k,
H_{<t}^k,
\mathrm{Sig}(K_o^k),
\mathrm{Ind}(K_p^k)
\right],
\]
where $\mathcal{I}$ is the benchmark instruction, $o_t^k$ is the new observation, and $H_{<t}^k$ is the trajectory so far. The object signatures expose public classes, methods, functions, and docstrings from $K_o^k$; the procedural index lists procedural files and their public functions. These summaries provide enough information for the agent to recognize potentially relevant knowledge without loading the entire knowledge base into context.

When the agent needs implementation details, it may issue an internal inspection request $q$ of the form \texttt{Inspect[...]} instead of an environment action. \method{} retrieves the matching object definitions or procedural files and expands the context:
\[
C_t^{(1)}
=
C_t^{(0)}
\oplus
\mathrm{Src}\!\left(
\mathrm{Retrieve}(q; K_o^k, K_p^k)
\right).
\]
Inspection is internal and bounded: if no matching code is found, the agent is informed that inspection failed; if the inspection budget is exhausted, it must produce an environment action.

Progressive disclosure prevents the growing knowledge model from overfilling the context window. Reflection and update organize experience into object classes and focused procedural files; signatures and indexes expose the high-level structure of that model; inspection reveals implementation details only when they are relevant to the present decision. Thus, \method{} can reuse an expanding body of learned knowledge while preserving the scalability, auditability, and consistency benefits of an executable object-centric environment model.

\section{Experiments}

\input{tables/main.tex}

\begin{figure*}[!t]
    \centering
    \includegraphics[width=\linewidth]{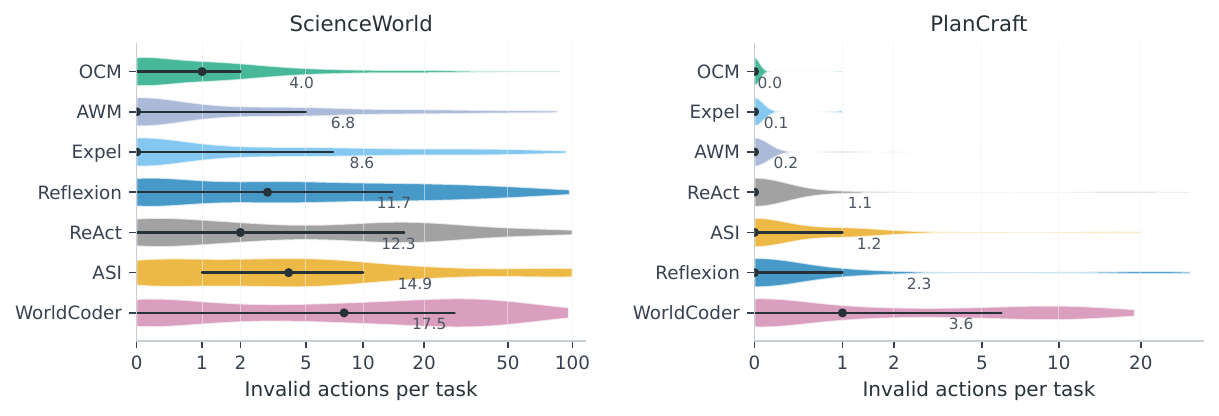}
    \caption{\textbf{Distribution of invalid actions across ScienceWorld and PlanCraft.} Each violin shows the per-task distribution of invalid actions for one method, with the x-axis plotted on a $\log(1+x)$ scale to reduce the effect of long-tailed failures. Black dots mark medians, horizontal black bars mark interquartile ranges, and numeric annotations report mean invalid actions per task in the original scale. OCM yields the lowest mean invalid-action count on both benchmarks, suggesting that object-centric environment modeling improves action feasibility.}
    \label{fig:invalid_actions}
    \vspace{-10pt}
\end{figure*}

\subsection{Experimental Setting}

\paragraph{Benchmarks.} We evaluate \method{} on three interactive agent benchmarks: ScienceWorld~\cite{wang2022scienceworld}, ALFWorld~\cite{shridhar2020alfworld}, and PlanCraft~\cite{dagan2024plancraft}, which respectively capture challenges in grounded scientific reasoning, embodied household interaction, and structured crafting dynamics. To ensure a fair comparison and control computational resources, we sampled five tasks from each of 30 ScienceWorld task categories as test data. Meanwhile, we used the unseen split for ALFWorld and the test (small) split for PlanCraft. More details are provided in Appendix~\ref{app:benchmarks}.

\noindent \paragraph{Baselines.} We compare \method{} with standard interactive agents, including ReAct~\cite{yao2022react} and Reflexion~\cite{shinn2023reflexion}; memory-augmented agents, including Expel~\cite{zhao2024expel} and AWM~\cite{wang2024agent}; and symbolic-learning or world-modeling agents, including ASI~\cite{wang2025inducing}, WorldCoder~\cite{tang2024worldcoder}, and Wall-E~\cite{zhou2024wall}. All methods are evaluated in the same online setting: agents process episodes sequentially and may update their memory or learned knowledge after each episode, but receive no offline training data or benchmark-specific demonstrations. We use GPT-4.1-mini as the backbone LLM~\cite{openai2025gpt41}. More implementation details are provided in Appendix~\ref{app:baselines}.

\noindent \paragraph{Metrics.} We report success rate (SR), average reward, average environment steps, and average rank (A.R.) across benchmark metrics. For PlanCraft, we additionally report results by difficulty level. To analyze interaction quality, we measure the number of invalid actions rejected by the environment.

\begin{figure*}[!t]
    \centering
    \includegraphics[width=\linewidth]{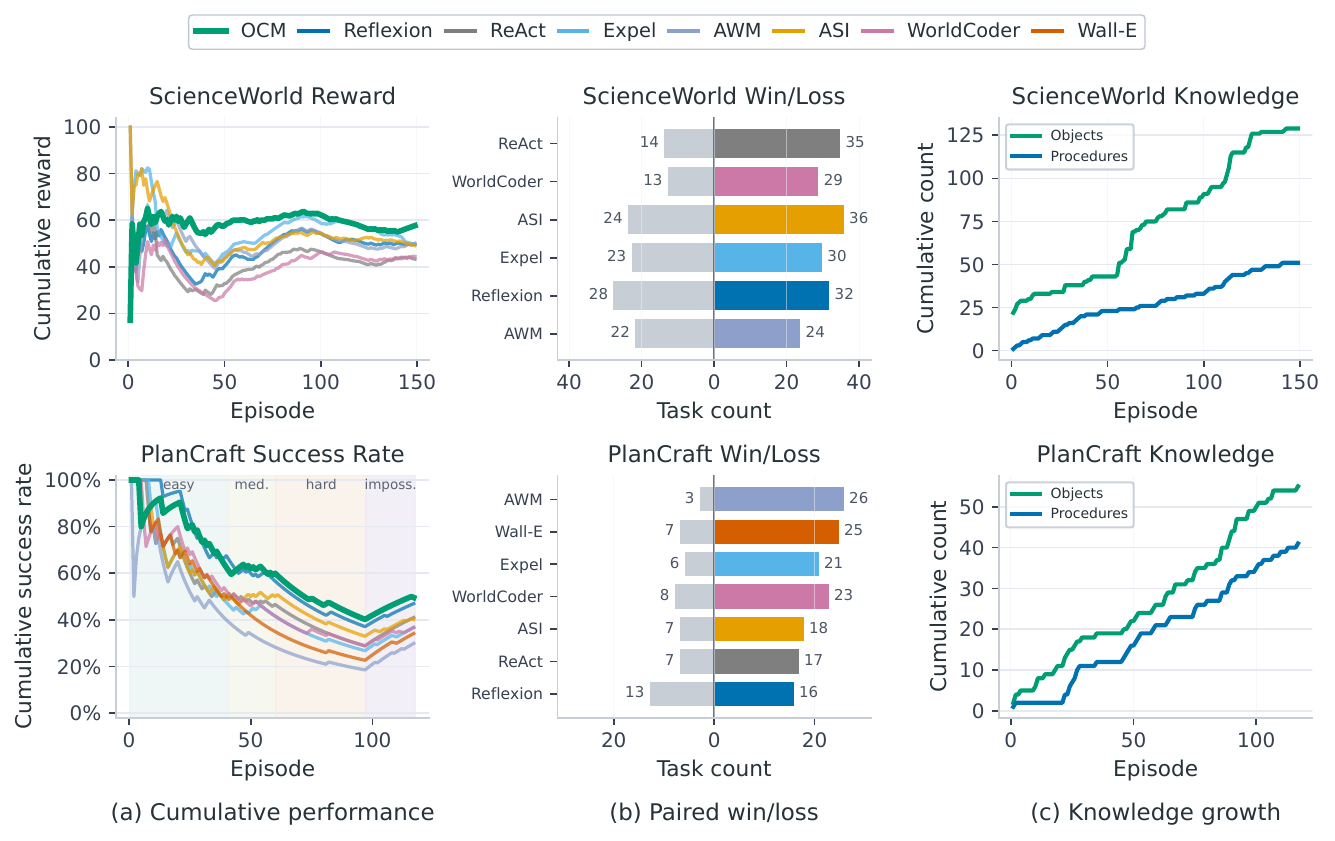}
    \caption{\textbf{Analysis of OCM across ScienceWorld and PlanCraft.} (a) Cumulative performance over the online evaluation sequence, measured by reward on ScienceWorld and success rate on PlanCraft. (b) Paired win/loss comparisons between OCM and each baseline on the same task instances. Bars to the right indicate tasks solved by OCM but not the baseline, while bars to the left indicate the opposite. (c) Growth of OCM's environment knowledge over time. Object and procedural entries both expand throughout evaluation, showing that OCM accumulates reusable environment structure rather than relying only on per-task reflection.}
    \label{fig:evolution}
    \vspace{-10pt}
\end{figure*}

\subsection{Main Results}

Table~\ref{tab:benchmark_results} shows that \method{} achieves the best average rank across the three benchmarks, indicating consistent gains across different environment families. The improvement is greatest on ScienceWorld and PlanCraft, where success depends heavily on reusable object mechanisms and state transformations. On ScienceWorld, \method{} obtains the highest success rate and reward while using a competitive number of steps. This suggests that object-centric knowledge helps the agent reuse recurring mechanisms such as containment, heating, cooling, and material transformation, rather than relying only on task-specific experience. PlanCraft provides a complementary test of structured transformation knowledge. \method{} achieves the best overall success rate, with especially clear gains on medium tasks. Easy tasks can often be solved by direct reasoning or short-horizon search, but harder recipes require tracking intermediate dependencies and avoiding invalid operations. The stronger gains in these settings are consistent with the design of \method{}, which can represent materials, tools, and transformations as reusable object-centered knowledge. Results on ALFWorld are more mixed: \method{} is competitive but does not outperform all baselines. This suggests that the benefits of object-centric modeling are most pronounced when transfer requires explicit mechanisms. In benchmarks dominated by high-level household routines, textual experiential lessons may already provide strong guidance.

\subsection{Learning Dynamics}

Figure~\ref{fig:evolution} examines how performance evolves over the online evaluation process. On ScienceWorld and PlanCraft, \method{} maintains a stronger cumulative performance curve over time, suggesting that its learned knowledge remains useful as more episodes are processed. This is important for online agents, where accumulated memory can easily become noisy, redundant, or overly specific. The paired win-loss comparison compares methods on the same task instances. Across both benchmarks, \method{} solves more tasks that each baseline fails to solve than vice versa. This indicates that the aggregate gains are not driven by a small number of outlier tasks, but reflect broader improvements across the evaluation set. The knowledge-growth curves further show that both object knowledge and procedure knowledge continue to expand during evaluation. This supports the intended division of labor: object knowledge captures reusable entities and mechanisms, while procedure knowledge records focused ways of applying them. 

\subsection{Action Validity Analysis}

Final success alone does not fully characterize agent behavior, since an agent may eventually solve a task after many invalid or unproductive actions. Figure~\ref{fig:invalid_actions} therefore measures invalid actions at the task level. \method{} has the lowest mean invalid-action count on both ScienceWorld and PlanCraft, indicating that its learned knowledge improves not only task completion but also action feasibility.

This result is closely aligned with the method design. Workflow-based memories can still fail when their implicit preconditions do not match the current state. \method{} reduces this risk by grounding procedural knowledge in shared object definitions and exposing object signatures during action selection. As a result, the agent can better reason about affordances and preconditions before acting.

\subsection{Ablation Study}

\begin{figure*}[!t]
    \centering
    \includegraphics[width=\linewidth]{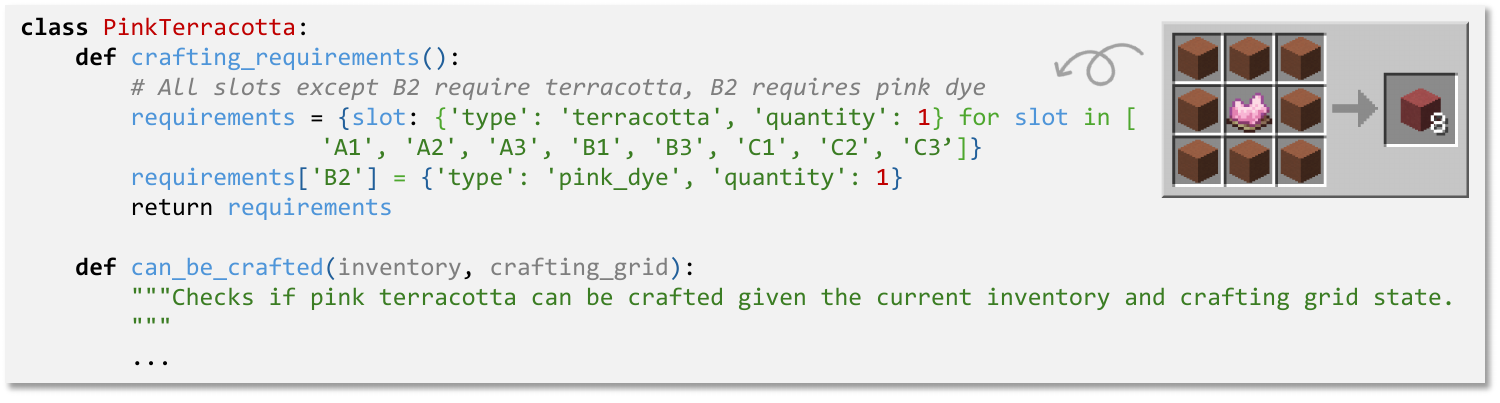}
    \caption{\textbf{Example of Object-Centric Knowledge Update.} From a successful PlanCraft trajectory for crafting \texttt{pink\_glazed\_terracotta}, OCM extracts object knowledge for the intermediate \texttt{pink\_terracotta} recipe.}
    \label{fig:case}
    \vspace{-10pt}
\end{figure*}

Table~\ref{tab:ablation_plancraft} isolates the contribution of the main components of \method{} on PlanCraft. \textit{Removing object knowledge} reduces the success rate, showing that procedural experience alone does not fully capture the reusable structure of crafting tasks. Without a shared object model, learned procedures are less reliably grounded in environment constraints. \textit{Removing procedure knowledge} also hurts performance and increases invalid actions. This indicates that object knowledge alone is not sufficient: object classes encode affordances and mechanisms, but the agent still benefits from procedure files that summarize recurring ways of applying them. The two knowledge bases therefore play complementary roles. The \textit{no-verification variant} suffers the largest drop in success rate. During online interactions, unverified updates can persist across future episodes and mislead later decisions. Verification helps prevent this by requiring newly written procedures to execute successfully against the current object knowledge before they are committed.

\input{tables/ablation.tex}

\subsection{Case Study}

We further examine how \method{} turns a successful interaction into reusable knowledge on a PlanCraft task whose goal is to craft \texttt{pink\_glazed\_terracotta}. The initial inventory contains \texttt{terracotta} and \texttt{pink\_dye}, but the target item is not directly craftable from these ingredients. The successful trajectory therefore has two stages: the agent first fills the crafting grid with \texttt{terracotta} in every slot except the center slot \texttt{B2}, places \texttt{pink\_dye} in \texttt{B2}, obtains \texttt{pink\_terracotta} from the output slot, and then smelts \texttt{pink\_terracotta} into \texttt{pink\_glazed\_terracotta}. This trajectory is useful because it contains both a recipe and a state transformation, rather than a single isolated action.

After the episode, \method{} writes object knowledge that makes the intermediate object explicit. As illustrated in Figure~\ref{fig:case}, the learned \texttt{PinkTerracotta} class defines \texttt{crafting\_requirements()}, which records the exact slot-level recipe, and \texttt{can\_be\_crafted()}, which checks whether the inventory and crafting grid satisfy those requirements without extra misplaced items. It also writes \texttt{PinkGlazedTerracotta}, which records that \texttt{pink\_terracotta} is the smeltable precursor. These object definitions separate stable environment mechanisms from the particular episode in which they were observed.

\method{} also writes procedure knowledge grounded in those objects. The \texttt{craft\_pink\_terracotta} procedure imports \texttt{PinkTerracotta}, uses its requirements to detect missing slots, wrong item types, insufficient quantities, and unexpected extra items, and records recovery principles for repairing the grid. The \texttt{smelt\_pink\_terracotta} procedure imports \texttt{PinkGlazedTerracotta} and encodes the precursor-to-output transformation. Both procedures include verifier tests, such as checking that a correct grid passes \texttt{can\_be\_crafted()}, and that a grid with an extra wrong item fails. This case illustrates the central benefit of object-centric modeling: the learned knowledge is not merely ``craft pink terracotta before smelting,'' but an executable account of the object constraints, failure modes, and transformation preconditions that make the experience reusable.

\section{Conclusion}

We presented \method{}, a framework for organizing interactive agent experience as an executable object-centric environment model. By grounding procedural knowledge in shared object definitions and verifying updates before committing them, \method{} supports more structured, reusable, and auditable experience accumulation than free-form memories or isolated skills. Experiments across three benchmarks show that \method{} improves overall performance and reduces invalid actions, suggesting that object-centric executable modeling is an effective substrate for online agent learning.

\section*{Acknowledgements}
This work was partially supported by the NSF under grants IIS-2533550, IIS-2321504, IIS-2217239, CNS-2426514, and CMMI-2146076, Notre Dame Strategic Framework Research Grant (2025), and Notre Dame Poverty Research Package (2025). Any expressed opinions, findings, and conclusions or recommendations are those of the authors and do not necessarily reflect the views of the sponsors.

\section*{Limitations}

OCM has several limitations that suggest directions for future work. 
First, our experiments focus on text-based interactive environments with relatively stable object types, action interfaces, and transition mechanisms. 
This setting is well aligned with object-centric modeling, but future work could study how to extend OCM to more open-ended, multimodal, or dynamically changing environments where object boundaries and affordances are less explicit.

Second, OCM verifies executable consistency rather than providing a formal guarantee of semantic correctness. 
The current verification step ensures that procedural knowledge remains runnable against the object knowledge base, which is useful for maintaining an auditable code model, but it does not prove that every learned mechanism is complete or universally correct. 
Future work could incorporate stronger automatically generated tests, counterexample search, or environment-grounded validation to further improve semantic reliability.

Third, OCM introduces additional computation for reflection, knowledge updating, inspection, and verification. 
Although progressive knowledge disclosure reduces the need to place the full accumulated code base into every prompt, the framework still requires extra LLM calls after episodes and occasional source-code inspection during interaction. 
Improving the efficiency of OCM is an important practical direction.

Finally, the quality of the learned object and procedure knowledge depends on the backbone model's ability to abstract reusable mechanisms from trajectories and write coherent code. 
Future work could evaluate how OCM scales across model families, model sizes, and more constrained code-generation settings.

\section*{Ethical Considerations}

This work evaluates LLM agents in controlled simulated environments and does not involve human subjects, user studies, personal data, or sensitive real-world applications. 
The proposed framework concerns the organization and verification of accumulated agent knowledge rather than deployment in high-stakes settings. 
We therefore do not identify direct ethical concerns beyond standard risks of LLM-based agents, such as imperfect reasoning or incorrect actions, which are limited here to benchmark environments.


\bibliography{custom}

\appendix

\section{Experimental Details}
\label{sec:appendix}

\subsection{Benchmark Details}
\label{app:benchmarks}

Table~\ref{tab:benchmark_details} summarizes the evaluation splits used in our experiments. ScienceWorld contains 30 task categories, and we sample up to five test instances from each category. The \texttt{identify-life-stages-2} category contains only four available instances under our filtering, resulting in 149 ScienceWorld tasks in total. For ALFWorld, we use the out-of-distribution evaluation split. For PlanCraft, we use the test-small split, which preserves the benchmark's difficulty structure while keeping online evaluation computationally tractable. In all benchmarks, episodes are evaluated in the benchmark's default task order without reordering, curriculum design, or task-order tuning.

\begin{table}[!ht]
\centering
\small
\setlength{\tabcolsep}{3.5pt}
\begin{tabular}{@{}llcc@{}}
\toprule
Benchmark & Evaluation split & \#Tasks & Step limit \\
\midrule
ScienceWorld & Sampled test & 149 & 100 \\
ALFWorld & OOD eval & 134 & 50 \\
PlanCraft & Test-small & 117 & 30 \\
\bottomrule
\end{tabular}
\caption{\label{tab:benchmark_details}Evaluation splits used in our experiments.}
\end{table}

PlanCraft overall results include easy, medium, hard, and impossible tasks. Impossible tasks are instances where the target item cannot be synthesized from the available inventory and recipes; on these tasks, the episode is counted as successful if the agent correctly uses the \texttt{impossible} action. PlanCraft exposes five high-level action types: \texttt{move}, \texttt{smelt}, \texttt{think}, \texttt{search}, and \texttt{impossible}. The \texttt{move} action transfers item stacks among inventory, crafting-grid, and output slots; \texttt{smelt} applies furnace-like transformations; \texttt{search} retrieves recipe information; and \texttt{impossible} lets the agent declare that the target item cannot be crafted from the available resources. The \texttt{think} action is handled specially: it is parsed by the runner as an internal reasoning action rather than submitted to the environment, so it is not counted as an environment step.

\subsection{Baseline Details}
\label{app:baselines}

ReAct~\cite{yao2022react} is a standard interactive baseline that alternates between a brief reasoning trace and a primitive environment action. Reflexion~\cite{shinn2023reflexion} augments this loop with a self-reflection after failed attempts. Since Reflexion does not maintain cross-task memory in our setting, we allow it one retry within the same task episode and keep at most three within-task reflections in context. ExpeL~\cite{zhao2024expel} maintains cross-task textual rules induced from prior trajectories; because this memory is reused across tasks, we evaluate ExpeL without an additional retry. ExpeL retrieves the two most similar prior experiences, keeps at most 20 learned rules, keeps the three most recent reflections in context, updates rules in batches of eight trajectories, critiques eight successful trajectories per update, and uses task-similarity retrieval with the \texttt{all-mpnet-base-v2} embedder. AWM~\cite{wang2024agent} stores successful action workflows and retrieves one workflow for later tasks, with one induction pass after each episode. ASI~\cite{wang2025inducing} induces symbolic task knowledge from experience. Wall-E~\cite{zhou2024wall} and WorldCoder~\cite{tang2024worldcoder} are included as symbolic-learning or world-modeling baselines. Wall-E uses one candidate action, mines rules every 100 trajectories, and allows up to three internal retries during rule generation. WorldCoder uses bandit planning with exploration rate 0.05, bandit constant 5.0, and at most three LLM planning steps. These hyperparameters match the important defaults used by the corresponding open-source implementations whenever applicable. For \method{}, we allow up to three attempts when parsing reflection plans or repairing generated knowledge updates.

The original WorldCoder formulation was not directly usable in our environments. These benchmarks are complex and partially observable, and the agent could not reliably learn effective transition and reward functions from raw interaction alone during decision making. To obtain a runnable and competitive adaptation, we let the agent synthesize a transition function and a reward function from the trajectory at the end of each episode. In later episodes, these learned functions are included in the context and used as auxiliary world-model information for action selection.

For fairness, all methods use the same backbone model and are evaluated in the same online order. We do not include benchmark examples, task-specific shortcuts, or hand-written benchmark-specific heuristics in any prompt. Prompts contain only generic interaction instructions and the action-format information needed to communicate with each environment.

\subsection{Prompt Templates}
\label{app:prompts}

\tcbset{appendixprompt/.style={enhanced,breakable,colback=white,colframe=black!55,colbacktitle=black!72,coltitle=white,boxrule=0.45pt,arc=1mm,left=4pt,right=4pt,top=4pt,bottom=4pt,fonttitle=\bfseries\small,listing only,listing options={basicstyle=\ttfamily\scriptsize,breaklines=true,breakatwhitespace=false,columns=fullflexible,keepspaces=true}}}

The following boxes reproduce the OCM prompt templates used in our implementation, with runtime values represented as placeholders.

\begin{tcblisting}{appendixprompt,title=OCM Action Prompt}
System:
You are an interactive environment agent with access to two learned Python resources. object_knowledge.py describes reusable environment object types, observable states, affordances, constraints, relations, and state changes. procedural_knowledge/ describes reusable interaction lessons: workflows, rules, failure patterns, insights, syntax constraints, decision heuristics, and recovery strategies that use those object types. These resources are for reading and reasoning, not for direct execution. Do not copy internal class, file, or function names as environment actions. Use signatures, procedural index entries, and any inspected code to choose the next real text action. Inspect relevant learned resources when you are uncertain about object affordances, action syntax, preconditions, success signals, failure causes, or recovery. Output exactly two lines: a brief Thought line and an Action line. Use internal inspect only to read learned resources, for example: Action: Inspect[object.Container, procedural.heating_substance]. Otherwise the Action line must be one primitive environment action.

Benchmark instruction:
{benchmark_instruction}

System:
object_knowledge.py signatures:
{object_signatures or "(empty)"}

procedural_knowledge/index.py:
```python
{procedural_index or "PROCEDURAL_KNOWLEDGE = {}"}
```

Inspected learned code available in this episode:
```python
{expanded_source}
```

Conversation history:
{previous user/assistant trajectory messages}

User:
{current_observation}
\end{tcblisting}

\begin{tcblisting}{appendixprompt,title=OCM Reflection Prompt}
System:
Analyze one completed environment episode and write a concise update plan for two learned Python resources that will be read during future decisions.

object_knowledge.py models reusable environment entities, states, affordances, constraints, relations, and state changes.
procedural_knowledge/*.py stores compact action-facing lessons that use object_knowledge types: workflows, rules, syntax constraints, failure patterns, decision heuristics, or recovery principles.

Design principles:
Separate the world model from action lessons. Ground every update in observations, invalid-action feedback, reward, done, or success signals. Treat existing knowledge as provisional: preserve it when supported, refine it when incomplete, and replace it when contradicted. Successful episodes show what works; failed episodes mostly show preconditions, invalid syntax, missing observations, failure causes, and recovery. Keep final knowledge compact and reusable.

Return JSON only. Do not return Markdown.

User JSON:
{
  "benchmark": "{benchmark_name}",
  "result": {result_meta},
  "object_knowledge_signatures": "{object_signatures}",
  "procedural_knowledge_index_py": "{procedural_index}",
  "episode_steps": {steps},
  "rules": {
    "output": [
      "Return JSON only with keys: object_knowledge_updates, procedural_knowledge_updates, inspect_objects, inspect_procedural.",
      "Each planned update should be compact: target, operation, lesson, evidence, reason.",
      "operation is create, update, or replace.",
      "inspect_objects contains object top-level names or ClassName.method_name names to inspect.",
      "inspect_procedural contains procedural file stems without .py to inspect."
    ],
    "selection": [
      "Plan only reusable knowledge supported by the episode.",
      "Use object updates for entities, states, affordances, constraints, relations, or mappings.",
      "Use procedural updates for workflows, rules, syntax constraints, failure patterns, heuristics, or recovery.",
      "Skip one-offs, exact locations, task ids, raw trajectories, and full action scripts."
    ],
    "content": [
      "Evidence should mention the relevant observation, invalid feedback, reward, done, or success signal.",
      "If existing knowledge is contradicted, plan replace rather than only adding new knowledge.",
      "For procedural updates, name the object_knowledge types or helpers the lesson should use."
    ]
  }
}
\end{tcblisting}

\begin{tcblisting}{appendixprompt,title=OCM Knowledge-Update Prompt}
System:
Implement updates from a completed environment episode for two learned Python resources that future decisions will read.

object_knowledge.py models reusable environment entities, states, affordances, constraints, relations, and state changes.
procedural_knowledge/*.py stores compact action-facing lessons that use object_knowledge types: workflows, rules, syntax constraints, failure patterns, decision heuristics, or recovery principles.

Design principles:
Separate the world model from action lessons. Ground every update in observations, invalid-action feedback, reward, done, or success signals. Treat existing knowledge as provisional: preserve it when supported, refine it when incomplete, and replace it when contradicted. Successful episodes show what works; failed episodes mostly show preconditions, invalid syntax, missing observations, failure causes, and recovery. Keep final knowledge compact and reusable.

Return JSON only with keys 'object_knowledge' and 'procedural_knowledge'. Each key maps to a list of objects with fields 'existing_name', 'name', and 'code'. For new knowledge, existing_name is empty. For replacement or rename, existing_name is the current object class name or procedural file stem. name is the resulting class name or file stem. code is the complete replacement code for that item.

User JSON:
{
  "object_knowledge_signatures": "{object_signatures}",
  "procedural_knowledge_index_py": "{procedural_index}",
  "update_plan": {update_plan},
  "inspected_source": "{inspected_source}",
  "result": {result_meta},
  "episode_steps": {steps},
  "previous_error": "{previous_error}",
  "rules": {
    "object_knowledge": [
      "Write compact classes or top-level helpers for reusable environment entities, states, affordances, constraints, relations, and mappings.",
      "Model only what the episode evidence supports.",
      "Use simple defaults so deterministic checks can instantiate classes without live environment access.",
      "Never import object_knowledge from inside object_knowledge.py.",
      "Do not write task solutions, action scripts, or episode transcripts."
    ],
    "procedural_knowledge": [
      "Each file represents one compact reusable interaction lesson.",
      "Import object_knowledge classes or helpers instead of redefining environment entities.",
      "Use short docstrings, small constants, typed records, and focused helpers.",
      "Action strings are allowed only as primitive syntax examples, not complete scripts or next-action APIs.",
      "Helpers should classify, explain, check applicability, or expose examples; they should not choose the next action.",
      "When replacing an existing file, remove claims contradicted by the episode."
    ],
    "quality": [
      "Use retrieval-friendly names and brief docstrings.",
      "Keep detailed evidence in the reflection plan, not in long code comments or docstrings.",
      "Every procedural file must include a lightweight no-argument __verify__() with deterministic assertions.",
      "__verify__ should check imports/references and simple helper behavior.",
      "__verify__ must not call the live environment or external services.",
      "The candidate knowledge must compile and import successfully."
    ]
  }
}
\end{tcblisting}

\begin{table*}[!t]
\centering
\small
\setlength{\tabcolsep}{5pt}
\begin{tabular}{@{}p{0.24\linewidth}p{0.23\linewidth}p{0.23\linewidth}p{0.21\linewidth}@{}}
\toprule
Stage & Evidence & OCM knowledge & Effect on \texttt{2-3\_225} \\
\midrule
Prior episodes & Related melting-point tasks expose repeated partial progress and failures on thermometer use, nested substance access, temperature/state reading, and threshold-based box selection. &
\method{} creates \textsc{MeltSort}, a reusable procedure with accessibility checks, temperature/state retrieval, threshold comparison, and conditional focus (Figures~\ref{fig:appendix_case_objects}--\ref{fig:appendix_case_code}). &
The relevant procedure is already available before the later unknown-substance task begins. \\
\midrule
Target episode & The task asks whether \texttt{solid unknown substance Y} has melting point above \texttt{200.0} degrees Celsius. All seven baselines fail on the same instance. &
During action selection, \method{} calls and inspects \textsc{MeltSort}, exposing the decision skeleton in Figure~\ref{fig:appendix_case_code} rather than a fixed action script. &
The agent decomposes the inspected source into legal actions, heats and re-measures the substance, observes \texttt{390} degrees Celsius, and focuses on the blue box. \\
\bottomrule
\end{tabular}
\caption{\label{tab:appendix_case_sciworld}Evidence that \method{} reuses previously learned procedural knowledge on a later ScienceWorld task where all baselines fail.}
\end{table*}

\subsection{Additional Case Study: Reusing Learned Procedural Knowledge}
\label{app:case-study}

We present an additional ScienceWorld case that illustrates how \method{} reuses knowledge accumulated from earlier episodes. Before episode \texttt{2-3\_225}, the agent had interacted with several related melting-point tasks. Those earlier trajectories did not all end in success, but they provided reusable evidence about the task structure: the agent must focus on a thermometer, access the target substance even when it is nested in a container, measure or induce its temperature, compare the result against a threshold, and then focus on the correct destination box. From these experiences, \method{} wrote a procedural entry for measuring and sorting a substance by melting point; we refer to this entry as \textsc{MeltSort} below. Figures~\ref{fig:appendix_case_objects} and~\ref{fig:appendix_case_code} show the object interface and procedural excerpt used by this case.

The later task \texttt{2-3\_225} asks the agent to measure the melting point of \texttt{solid unknown substance Y} and focus on the blue box if the measured temperature is above \texttt{200.0} degrees Celsius, otherwise on the orange box. This task is difficult for baselines: ReAct, Reflexion, ExpeL, AWM, ASI, WorldCoder, and Wall-E all fail on the same instance, while \method{} succeeds with reward 100. During decision making, \method{} calls the learned melting-point procedure and inspects its source. The procedure itself is not an environment action, so the agent then converts the inspected procedure into valid ScienceWorld actions: it focuses on the thermometer and substance, picks up the thermometer, uses it on the substance, opens the oven, moves the substance into the oven, activates the oven, waits, measures again, observes a temperature of \texttt{390} degrees Celsius, and focuses on the blue box. The case shows that the learned procedure is useful not as a script to copy verbatim, but as an executable and inspectable model of the hidden task logic.

\begin{figure*}[!t]
\centering
\begin{tcblisting}{enhanced,colback=white,colframe=black!55,colbacktitle=black!72,coltitle=white,boxrule=0.45pt,arc=1mm,left=3pt,right=3pt,top=3pt,bottom=3pt,fonttitle=\bfseries\small,listing only,listing options={basicstyle=\ttfamily\scriptsize,breaklines=true,breakatwhitespace=false,columns=fullflexible,keepspaces=true},title=Object Knowledge Interface Used by the Procedure}
class ObjectKnowledge:
    def __init__(self, name, location=None):
        self.name = name
        self.location = location
        self.contains = []
        self.properties = {}

    def get_state(self):
        return self.properties.get("state")

    def get_temperature(self):
        return self.properties.get("temperature")

    def is_living_thing(self):
        return False

    def is_in_inventory(self):
        return self.location is None

    def focus(self):
        if self.is_container():
            living = self.get_living_things()
            if living:
                return living[0]
            substances = [
                obj for obj in self.contains
                if not obj.is_living_thing() and not obj.is_container()
            ]
            if len(substances) == 1:
                return substances[0]
        return self

class Container(ObjectKnowledge):
    def is_accessible(self):
        return self.properties.get("is_open", True)

    def can_be_opened(self):
        return self.properties.get("openable", False)

    def can_accept(self, obj):
        if not self.is_accessible():
            return False
        capacity = self.properties.get("capacity")
        if capacity is not None and len(self.contains) >= capacity:
            return False
        return True

    def open_container(self):
        if self.can_be_opened():
            self.properties["is_open"] = True
        return self.is_accessible()
\end{tcblisting}
\caption{\label{fig:appendix_case_objects}Excerpt of the object knowledge used by the melting-point procedure. The full \texttt{object\_knowledge.py} contains additional unrelated methods; here we retain the state, temperature, focus, inventory, and container-accessibility interfaces needed for this case.}
\end{figure*}

\begin{figure*}[!t]
\centering
\begin{tcblisting}{enhanced,colback=white,colframe=black!55,colbacktitle=black!72,coltitle=white,boxrule=0.45pt,arc=1mm,left=3pt,right=3pt,top=3pt,bottom=3pt,fonttitle=\bfseries\small,listing only,listing options={basicstyle=\ttfamily\tiny,breaklines=true,breakatwhitespace=false,columns=fullflexible,keepspaces=true},title=Learned Procedural Knowledge}
from typing import Optional
from object_knowledge import ObjectKnowledge, Container

def is_container_accessible(container: Container) -> bool:
    if container.is_accessible():
        return True
    if container.can_be_opened():
        container.open_container()
        return container.is_accessible()
    return False

def get_substance_temperature_or_state(
    substance: ObjectKnowledge
) -> Optional[float]:
    if substance.is_living_thing():
        return None
    state = substance.get_state()
    if state is not None and state.lower() == "air":
        return None

    temperature = substance.get_temperature()
    if temperature is not None:
        return temperature

    if state is not None:
        if "molten" in state.lower() or "liquid" in state.lower():
            return 250.0
        elif "solid" in state.lower() or "frozen" in state.lower():
            return 10.0
    return None

def focus_on_object(obj: ObjectKnowledge) -> bool:
    state = obj.get_state()
    if state is not None and state.lower() == "air":
        return False
    if obj.is_in_inventory():
        return True
    focused = obj.focus()
    return focused is not None

def use_thermometer_on_substance(
    thermometer: ObjectKnowledge,
    substance: ObjectKnowledge,
) -> Optional[float]:
    temp = thermometer.get_temperature()
    if temp is not None:
        return temp
    return get_substance_temperature_or_state(substance)

def measure_and_sort_substance_by_melting_point_with_nested_container_handling(
    thermometer: ObjectKnowledge,
    substance: ObjectKnowledge,
    container_if_above_threshold: Container,
    container_if_below_threshold: Container,
    melting_point_threshold: float = 10.0,
) -> bool:
    if not focus_on_object(thermometer):
        return False
    if not focus_on_object(substance):
        return False

    temperature = use_thermometer_on_substance(thermometer, substance)
    if temperature is None:
        return False

    if temperature > melting_point_threshold:
        target_container = container_if_above_threshold
    else:
        target_container = container_if_below_threshold

    if not is_container_accessible(target_container):
        return False
    if not focus_on_object(target_container):
        return False

    return True
\end{tcblisting}
\caption{\label{fig:appendix_case_code}Excerpt of the learned procedural code used by \method{} for measuring and sorting a substance by melting point. The full procedure contains longer docstrings and additional comments; this excerpt preserves the executable control flow and calls into the object interface in Figure~\ref{fig:appendix_case_objects}.}
\end{figure*}

In \texttt{2-3\_225}, the procedure arguments bind to the thermometer, \texttt{solid unknown substance Y}, the blue box, the orange box, and threshold \texttt{200.0}. The inspected code therefore gives the agent the correct decision skeleton, while the online trajectory supplies the concrete environment actions needed to realize it.

\end{document}

%% file: tables/main.tex
\begin{table*}[!t]
\centering
\resizebox{\linewidth}{!}{
\begin{tabular}{c|ccc|>{\centering}p{1.2cm}>{\centering}p{1.2cm}|ccccc|c}
\toprule
\multirow{2}{*}{\textbf{Method}}
& \multicolumn{3}{c|}{\textbf{ScienceWorld}}
& \multicolumn{2}{c|}{\textbf{ALFWorld}}
& \multicolumn{5}{c|}{\textbf{PlanCraft}}
& \multirow{2}{*}{\textbf{A.R.}} \\
\cmidrule{2-11} 
& \textbf{SR} & \textbf{Reward} & \textbf{Steps}
& \textbf{SR} & \textbf{Steps}
& \textbf{Easy} & \textbf{Med.} & \textbf{Hard} & \textbf{Overall} & \textbf{Steps}
&  \\
\midrule
ReAct
& 20.8 & 43.3 & 46.7
& 29.9 & 39.8
& 50.0 & 40.0 & 0.0 & 41.0 & \textbf{7.3}
& 4.90 \\

Reflexion
& 32.2 & 49.6 & 50.9
& 30.6 & \textbf{17.2}
& \textbf{67.5} & 35.0 & \underline{5.4} & \underline{47.0} & \underline{8.4}
& \underline{2.95} \\

Expel
& 30.2 & \underline{52.3} & 40.6
& \textbf{60.1} & \underline{30.6}
& 47.5 & 30.0 & 2.7 & 36.8 & 13.1
& 4.15 \\

AWM
& \underline{33.6} & 50.1 & \textbf{39.0}
& 37.3 & 40.4
& 40.0 & 5.0 & 2.7 & 29.9 & 12.9
& 4.85 \\

ASI
& 26.9 & 49.1 & 52.8
& 10.5 & 47.0
& 50.0 & \underline{50.0} & \underline{5.4} & 40.2 & 12.7
& 5.15 \\

WorldCoder
& 24.2 & 44.4 & 53.5
& 25.4 & 43.1
& 52.5 & 20.0 & \textbf{8.1} & 36.8 & 16.7
& 5.70 \\

Wall-E
& 23.7 & 44.1 & 52.3
& 27.6 & 42.7
& 50.0 & 5.0 & 2.7 & 34.2 & 17.2
& 6.55 \\

\midrule
\textbf{OCM}
& \textbf{34.8} & \textbf{57.7} & \underline{40.1}
& \underline{41.7} & 34.5
& \underline{62.5} & \textbf{55.0} & \textbf{8.1} & \textbf{49.6} & 8.8
& \textbf{1.75} \\
\bottomrule
\end{tabular}
}
\caption{\textbf{Performance and Average Ranking Comparison.} Results are reported on ScienceWorld, ALFWorld, and PlanCraft. \textbf{SR} means Success Rate. The best and second-best performances are highlighted in \textbf{Bold} and \underline{Underline}.}
\label{tab:benchmark_results}
\end{table*}

%% file: tables/ablation.tex
\begin{table}[t]
\centering
\resizebox{0.9\columnwidth}{!}{
\begin{tabular}{l|>{\centering}p{1.2cm}cc}
\toprule
\multirow{2}{*}{\textbf{Variants}}
& \multicolumn{3}{c}{\textbf{PlanCraft}} \\
\cmidrule{2-4}
& \textbf{SR} & \textbf{Steps} & \textbf{Invalid} \\
\midrule
\textbf{OCM}
& \textbf{49.6} & \underline{8.8} & \textbf{3} \\

-- w/o $K_o$
& 44.4 & 9.5 & 6 \\

-- w/o $K_p$
& \underline{45.3} & 9.7 & 33 \\

-- w/o verification
& 39.3 & \textbf{6.5} & \underline{4} \\
\bottomrule
\end{tabular}
}
\caption{Ablation results of OCM on PlanCraft.}
\label{tab:ablation_plancraft}
\vspace{-10pt}
\end{table}